\documentclass[review]{elsarticle}

\pdfoutput=1

\usepackage{lineno,hyperref}
\modulolinenumbers[5]

\usepackage{amsmath}

\usepackage{algorithmic}

\usepackage{array}
\usepackage[table]{xcolor}

\usepackage{dblfloatfix}
\usepackage{multirow}

\usepackage{epsfig}
\usepackage{calc}
\usepackage{amssymb}
\usepackage{amstext}
\usepackage{amsthm}
\usepackage{mathtools}
\usepackage{multicol}
\usepackage{pslatex}
\usepackage{apalike}
\usepackage{textcomp}
\usepackage[utf8]{inputenc}

\usepackage{tikz}

\usepackage{pifont}

\DeclarePairedDelimiterX{\norm}[1]{\lVert}{\rVert}{#1}

\journal{Pattern Recognition}

\begin{document}

\begin{frontmatter}

\title{Advanced local motion patterns for macro and micro facial expression recognition}

\author[mymainaddress]{B. Allaert\corref{mycorrespondingauthor}}
\ead{benjamin.allaert@univ-lille1.fr}
\author[mymainaddress]{IM. Bilasco}
\ead{marius.bilasco@univ-lille1.fr}
\author[mymainaddress]{C. Djeraba}
\cortext[mycorrespondingauthor]{Corresponding author}
\ead{chabane.djeraba@univ-lille1.fr}

\address[mymainaddress]{Univ. Lille, CNRS, Centrale Lille, UMR 9189 - CRIStAL - \\ Centre de Recherche en Informatique Signal et Automatique de Lille, F-59000 Lille, France}

\begin{abstract}

In this paper, we develop a new method that recognizes facial expressions, on the basis of an innovative local motion patterns feature, with three main contributions. The first one is the analysis of the face skin temporal elasticity and face deformations during expression. The second one is a unified approach for both macro and micro expression recognition. And, the third one is the step forward towards in-the-wild expression recognition, dealing with challenges such as various intensity and various expression activation patterns, illumination variation and small head pose variations. Our method outperforms state-of-the-art methods for micro expression recognition and positions itself among top-rank state-of-the-art methods for macro expression recognition.

\end{abstract}


\end{frontmatter}


\section{Introduction}
\label{sec:introduction}

Facial expression recognition has attracted great interest over the past decade in wide application areas, such as human machine interaction, behavior analysis, video communication, face identification/detection, face recognition, e-learning, well-being, e-health and marketing. For example, during visio-conferences between several participants, facial expression analysis strengthens dialogue and social interaction between all participants (i.e keep the viewers attention). In e-health applications, facial expressions recognition helps to better understand patient minds and pain, without any intrusive sensors.

\begin{figure}[!h]
\centering
\includegraphics[width=0.7\textwidth]{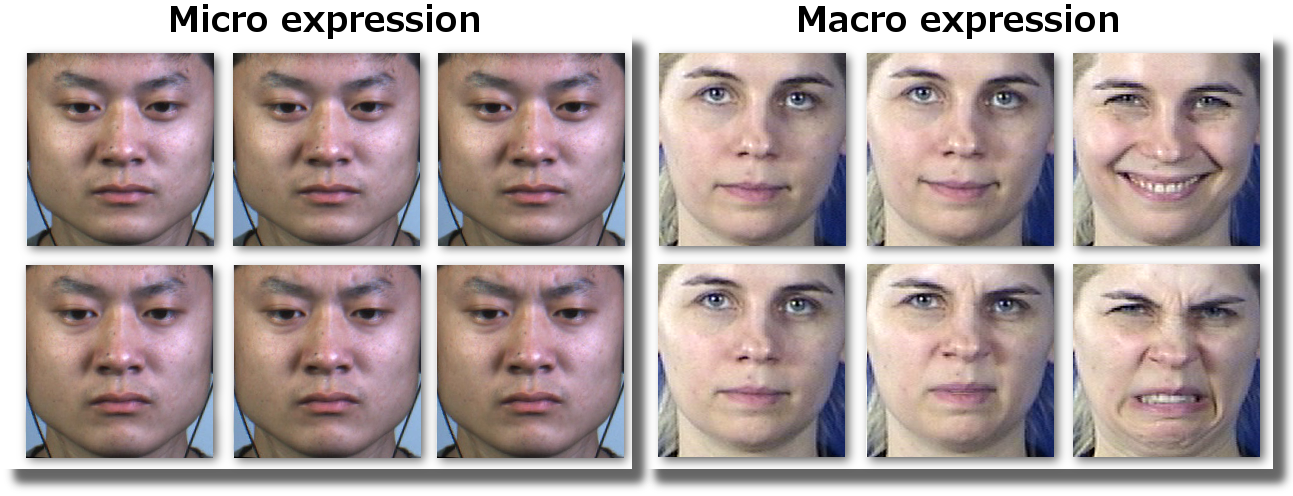}
\caption{Difference of motion intensity between micro and macro expression - happiness (line1), disgust (line2), from CASME II \cite{yan2014casme} and MMI \cite{pantic2005web}, respectively micro and macro expression datasets.}
\label{fig:magnify}
\end{figure}

Facial expressions are fundamentally covering both macro and micro expressions\cite{ekman1997face}. It is a very important issue, because by essence, both macro and micro expressions are present during human interactions. For example “happiness” expression ( Figure \ref{fig:magnify}) may be present in the form of micro or macro during conversations. So, dealing with both expression categories, in a unified approach, is meeting an important in-the-wild requirement.

The difference between macro and micro expression depends essentially of the duration and the intensity of expression, as illustrated in Figure \ref{fig:magnify}.

Macro expressions are voluntary facial expressions, and cover large face area. The underlying facial movements and the induced texture deformations can be clearly discriminated from the noises. The typical duration of macro expression is between 0.5 and 4 s \cite{ekman1997face}. Macro expressions are also characterized by high intensities, in terms of facial muscles movements and texture changes. So motion propagation is continued in facial area. On the opposite, micro expressions are involuntary facial expressions. Often, they convey hidden emotions that determine true human feelings and state-of-mind. Micro expressions tend to be subtle manifestations of a concealed emotion under a masked expression. Micro expressions are characterized by rapid facial movements and cover restricted / fragment facial area. The typical duration of micro expressions is between 65 ms and 500 ms \cite{yan2013fast}. Micro expressions are also characterized by low intensities \cite{porter2008reading,yan2014micro}, in terms of facial muscles movement and texture changes.

We propose a new method with innovative motion descriptor called local motion patterns (LMP), with three main contributions.
First, it takes into account mechanical facial skin deformation properties (local coherency and local propagation). Local motion patterns feature is spatio-temporal, and it filters discontinuities of motion adapted for low and high amplitudes.
Second, the method is a unified approach for micro expressions (disgust, happiness, repression, surprise) and macro expressions (anger, disgust, fear, happiness, sadness, surprise) recognition. When extracting motion information from the face, the unified approach deals with inconsistencies and noise, caused by face characteristics (skin smoothness, skin reflect and elasticity). 
Generally, related works on facial expression recognition have been proposed to deal separately with macro and micro expressions. And, very few of them consider both \cite{li2015reading}, because, especially for micro expression, the true facial motion is not discriminated from the motion noise caused by face characteristics. Third, on the basis of local facial motion intensity and propagation, the method is the step forward and potentially suitable for in-the-wild expression recognition: filtering out noise, showing robustness to illumination variation (near infrared and natural illumination), supporting small head pose variations, enhancing partial expression recognition and dealing with all expressions that lead to facial skin deformation.

Our face expression recognition method is validated on representative datasets of facial expression recognition community for both micro (CASME II, SMIC) and macro expression (CK+, Oulu-CASIA, MMI) analysis. The performances of our method are higher than state of the art methods for micro expression recognition, and are competitive compared to state of the art methods for macro expression recognition, considering only the initial data (without any augmentation).

\begin{figure}[!h]
\centering
\includegraphics[width=0.8\textwidth]{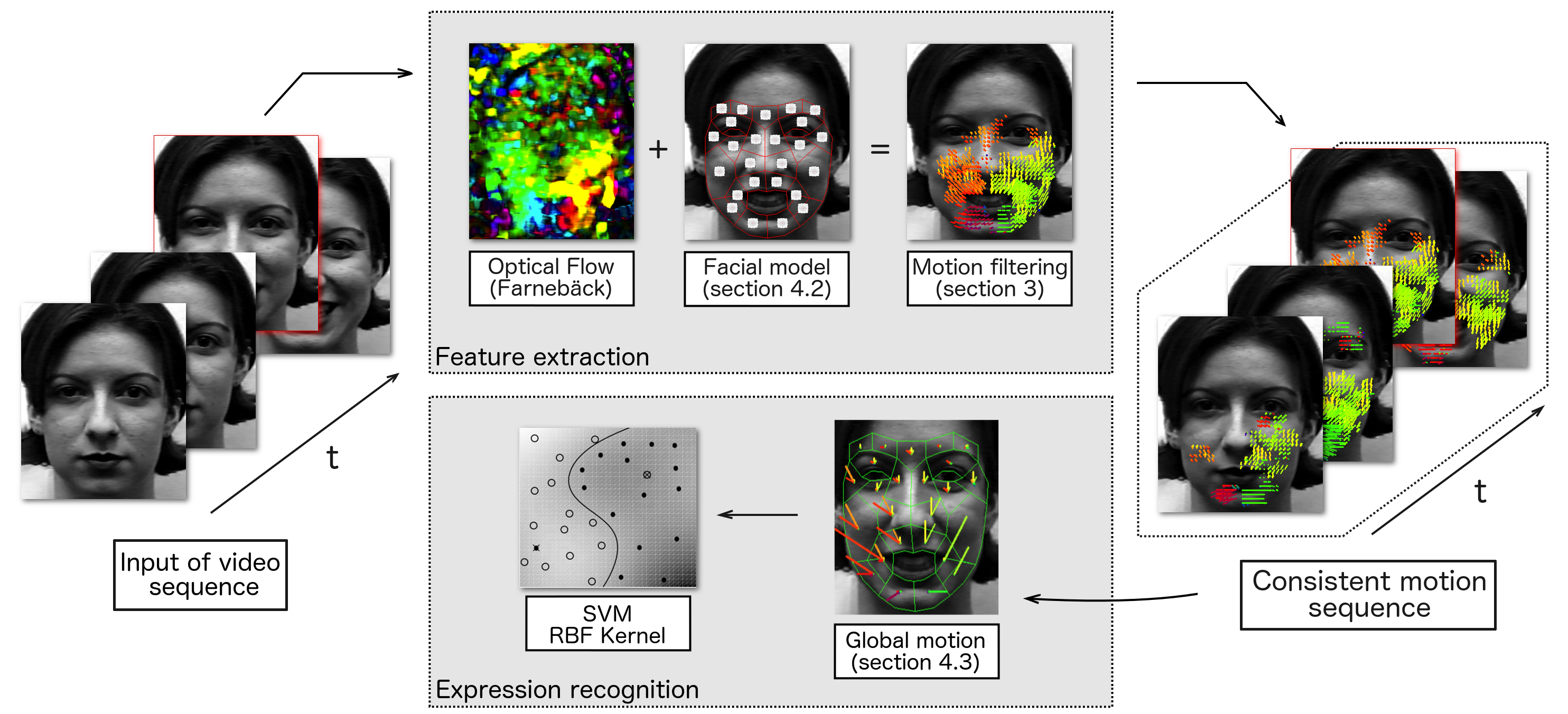}
\caption{Overview of our expression recognition method.}
\label{fig:schemeT1}
\end{figure}

In section \ref{sec:related}, we discuss works related to expression recognition. We introduce facial expression features, and current approaches, including handcrafted and deep-learning methods. 
In section \ref{sec:filtrage}, we present our local motion patterns feature that considers local motion coherency of the face (see "Feature extraction" part in Figure \ref{fig:schemeT1}). In this section, we show how local motion patterns feature deals with facial skin smoothness, reflection and elasticity. In section \ref{sec:reconnaissance}, we explore several strategies of encoding the facial motion for macro and micro expression recognition (see "Expression recognition" part in Figure \ref{fig:schemeT1}). We show the impact of the facial framework on recognition performances. Experimental results, presented in section \ref{sec:evaluation}, outline the generality of our method for micro and macro expression recognition
Conclusions, summing up the main contributions, and perspectives are discussed in section \ref{sec:conclusion}.

\section{Related Work}
\label{sec:related}

This section presents the most significant macro and micro facial expression recognition (FER) approaches that have been proposed in the literature. FER approaches are based on features and facial frameworks. The facial framework defines the regions of the faces, considering different granularities from whole face to points, where information is extracted. The information is composed of features, encoding changes of texture and motion. We start the section by discussing features of macro and micro expression recognition, followed by facial frameworks. Finally, We focus on the combination of features and facial frameworks for macro and micro expression recognition.

\subsection{Macro expression recognition}

Important motions induced by face skin muscles characterize macro expressions. Furthermore, with regard to facial deformation analysis, several types of techniques based on appearance and/or geometry can be distinguished to encode the changes.

Features, such as LBP \cite{ojala2002multiresolution} or HOG \cite{khan2012human} obtained good results in the analysis of macro facial deformations. A similar comment applies to convolutional neural network (CNN) approaches \cite{lopes2017facial,ding2017facenet2expnet,mollahosseini2016going}, where they learn a spatial feature representation on the apex frames 
By relying on the spatial feature only, LBP, HOG and static CNN approaches do not utilize facial expression dynamics while performing the facial expression recognition task, which can limit the performance
in presence of subtle expressions.

Psychological experiments by Bassili \cite{bassili1979emotion} showed that facial expressions could be recognized more accurately in a sequence of images. Therefore, a dynamic extension of LBP called Local Binary Pattern on Three Orthogonal Plans (LBP-TOP) is proposed by \cite{zhao2007dynamic}. Considering the latest developments in dynamic texture domains, the optical flow have regained interest from the community becoming one of the most widely used and recognized solution \cite{liao2013learning, fortun2015optical}. Optical flow methods are popular, because they estimate in a natural way the local dynamics and temporal texture characteristics. In recent deep learning approaches \cite{breuer2017deep,zhang2017facial}, a recurrent neural network (RNN) was used with the conventional CNN to encode dynamics in the sequence for classification of facial expression, and showed that the architectures of CNN with RNN can improve recognition performances compared to conventional CNN.

Most geometric feature-based approaches use Active Appearance Model (AAM) or variations like Active Shape Model (ASM), to track a dense set of facial points \cite{ghimire2013geometric,jung2015joint}. The locations of these facial landmarks are then used in different ways to extract the shape of facial features, and motion of facial features \cite{majumder2014emotion}.

The hybrid approaches combine geometric and appearance features. As suggested in \cite{kotsia2008texture}, combining features provides additional information to the recognition process. Han et al. \cite{han2014facial} use an AAM to 
improve feature extraction for recognizing facial Action Units (AUs). Jaiswal et al. \cite{jaiswal2016deep} use a combination of Convolutional and Bi-directional Long Short-Term Memory Neural Networks (CNN-BLSTM), which jointly learn shape,appearance and dynamics.
They show that the combination of dynamic CNN features and BLSTM excels at modeling the temporal information. Several deep learning methods \cite{jung2015joint,zhang2017facial} used a temporal geometric feature in order to reduce the effect of the identity on the learned spatio-temporal appearance features.


\subsection{Micro expression recognition}

Expression recognition approaches presented above, designed for macro expression, are not adapted to micro expression challenges (very short duration, low motion amplitude and texture changes). Liu et al. \cite{liu2015main} apply directly macro expressions approaches to micro expressions and show that detecting subtle changes
by applying traditional macro expression approaches is a difficult task. Indeed, partial and low-intensity facial movements in micro expressions differ from ordinary expressions and it is difficult to split between true facial motion and noise due to head movement or motion discontinuities. The same conclusion has been drawn when using deep learning \cite{patel2016selective}.

According to \cite{li2015reading}, micro expressions are much more difficult to detect without temporal information. Thus, researchers use spatio-temporal features for micro expression analysis. Liong et al. \cite{liong2014subtle} extend LBP-TOP using optical strain information as a weighting function to find smaller motions. Wang et al. \cite{wang2014lbp} propose an extension of LBP-TOP based on the three intersecting lines crossing over the center point of the three histograms. They provide more compact and lightweight representation by minimizing the redundancy in LBP-TOP. Huang et al. \cite{huang2016spontaneous2} propose a new spatio-temporal LBP on an improved integral projection combining the benefit of texture and shape.

Although most micro expression recognition studies have considered LBP-TOP, several authors investigate alternative methods. Huang et al. \cite{huang2016spontaneous} propose spatio-temporal completed local quantized pattern (STCLQP) and obtain promising performances. The reason may be that STCLQP provides more useful information for micro expression recognition, as it extracts jointly information characterizing magnitudes and orientations. Li et al. \cite{li2015reading} employ temporal interpolation and motion magnification to counteract the low intensity of micro expressions. They show that longer interpolated sequences do not lead to improve performances, because the movement tends to be diluted, and interpolating micro expression segments using only 10 frames is enough. Recently, Liu et al. \cite{liu2015main} design a feature for micro expression recognition based on a robust optical flow method, and extract Main Directional Mean Optical-flow (MDMO). They showed that the magnitude is more discriminant than the direction when working with micro expression.
Furthermore, several deep-learning methods have been proposed to deal with micro expression \cite{breuer2017deep,kim2017multi,patel2016selective} and they all present low performances.

In this context, systems based on dynamic textures provide better performance. Indeed, they allow detecting subtle changes occurring on the face, and do not require large changes in appearance, as texture-based or geometry-based approaches do. 

\subsection{Facial Frameworks}

The facial frameworks, based on geometric information, extract the most appropriate facial expression features. Assuming the face regions are well aligned; histogram-like features are often computed from equal-sized facial grids \cite{fan2017dynamic}. However, apparent misalignment can be observed, primarily caused by facial deformation of the expression itself. In most cases, the geometric features are used to ensure that facial regions (eyes, eyebrows, lip corners) are well matched with facial frameworks.

Appearance features extracted from active face regions improve the performance of expression recognition. Therefore, some approaches define the regions with respect to facial locations (i.e. eyes, lip corners) using geometrical characteristics of the face \cite{happy2015automatic}.

Recent studies use facial landmarks to define facial regions. They increase robustness to facial deformation analysis during expression. Jiang et al. \cite{jiang2014decision} define a mesh over the whole face with ASM, and extract features from the regions enclosed by the mesh. Sadeghi et al. \cite{sadeghi2013facial} used fixed geometric model for geometric normalization of facial images. The face image is divided into small sub-regions and then LBP histograms are calculated in each one for accurately describing the texture.

Facial frameworks based on salient patches, blocks, meshes or weighted masks have been explored overtime in combination with various features. However, despite the use of similar features in macro and micro expression recognition, it is still difficult to find a unified facial framework for analyzing macro and micro expression together.

\subsection{Discussions / Synthesis}

Micro expressions are quite different from macro expressions in terms of facial motion amplitudes and texture changes, which make them more difficult to characterize. Results from significant state-of-the-art approaches are illustrated in Table \ref{synthese}. The table shows the striking difference between macro and micro expression performances.

Table \ref{synthese} illustrates the established trends: appearance (static approaches), geometry and motion (dynamic texture and temporal approaches) in both macro and micro expression recognition fields. The main focus of the table resides in the difference in terms of performances between micro and macro expression recognition when the same underlying features and facial frameworks are used. Macro and micro expression recognition approaches are not directly comparable due to the fact that the underlying data is very different. However, we present them together in order to show that methods working well in one situation do not provide equivalent performances in the other. In order to allow an intra-category ranking, all macro expression approaches, cited in Table \ref{synthese}, use SVM as a final classifier and 10 fold cross-validation protocol, and all cited micro expression approaches use leave-one-subject-out cross validation protocol.

\begin{table}[!h]
\caption{State-of-the-art approaches on macro and micro expressions (* Data augmentation / Deep learning).}
\label{synthese}
\centering
\fontsize{7}{6}\selectfont
\begin{tabular}{ccccc}
\hline
Based on & \multicolumn{2}{c}{Macro expression (CK+)} & \multicolumn{2}{c}{Micro expression (CASME II)} \\
\hline
\multirow{6}*{Appearance} & LBP \cite{shan2009facial} & \multirow{2}*{90.05\%} & LBP \cite{li2015reading} & \multirow{2}*{55.87\%} \\
& Block-based & & Block-based & \\
\cline{2-5}
& PHOG \cite{khan2012human} & \multirow{2}*{95.30\%} & HIGO \cite{li2015reading} & 67.21\% \\
& Salient region & & Block-based & magnified \\
\cline{2-5}
& CNN \cite{lopes2017facial} & \multirow{2}*{* 96.76\%} & CNN \cite{patel2016selective} & \multirow{2}*{* 47.30\%} \\
& Whole face & & Whole face & \\
\hline
\multirow{4}*{Geometry} & Gabor Jet \cite{ghimire2013geometric} & \multirow{2}*{95.17\%} & \multirow{2}*{/} & \multirow{2}*{/} \\
& Facial points & & & \\
\cline{2-5}
& DTGN \cite{jung2015joint} & \multirow{2}*{* 92.35\%} & \multirow{2}*{/} & \multirow{2}*{/} \\
& Facial points & & & \\
\hline
\multirow{6}*{Motion} & LBP-TOP \cite{zhao2007dynamic} & \multirow{2}*{\textbf{96.26\%}} & DiSTLBP-IIP \cite{huang2016spontaneous2} & \multirow{2}*{64.78\%} \\
& Block-based & & Block-based & \\
\cline{2-5}
& Optical flow \cite{allaert2017maps} & \multirow{2}*{93.17\%} & MDMO \cite{liu2015main} & \multirow{2}*{\textbf{67.37\%}} \\
& Facial meshes & & Facial meshes & \\
\cline{2-5}
& CNN + AUs + LSTM \cite{breuer2017deep} & \multirow{2}*{\textbf{* 98.62\%}} & CNN + LSTM \cite{kim2017multi} & \multirow{2}*{* 60.98\%} \\
& Whole face & & Whole face & \\
\hline
\end{tabular}
\end{table}

As shown in Table \ref{synthese}, well-known static methods like LBP have limited potential for micro expression recognition. The difference would be attributable to the fact that it cannot discriminate very low intensity motions \cite{li2015reading}. LBP-TOP has shown promising performance for facial expression recognition. Therefore, many researchers have actively focused on the potential ability of LBP-TOP for micro expression recognition.

The geometric-based approaches deliver good results for some facial motions in macro expression situations, but fail in detecting subtle motions in presence of micro expressions. Subtle motions requires measuring skin surface changes. Algorithms tracking landmarks do not deliver the necessary accuracy for micro expression.

Dynamic texture approaches are best suited to low facial motion amplitudes \cite{huang2016spontaneous2}. Specifically, methods based on optical flow appear to be promising for micro expression analysis \cite{liu2015main}. Moreover, optical flow approach obtains competitive results in both macro and micro expression analysis \cite{allaert2017maps}. However, the optical flow approaches are often criticized for being heavily impacted by the presence of motion discontinuities and illumination changes. Recent optical flow algorithms (i.e. \cite{revaud2015epicflow}) evolved to better deal with noise. The majority of these algorithms is based on complex filtering and smooth propagation of motion to reduce the discontinuity of local motion, improving the quality of optical flow. Still, in presence of high and low intensity of motion, the smoothing effect tends to induce false motion. Another technique consists of artificially amplifying the motion. This technique is being used increasingly and successfully in micro expression recognition \cite{li2015reading}. The main disadvantage is the requirement of high intensity facial deformation in order to be efficient. Such deformations alter significantly the facial morphology, especially in the presence of macro expression.

Concerning deep learning approaches, we underline important contrasts. On one hand, deep learning approaches provide good results for macro expression recognition (see * lines in Table \ref{synthese}). Deep learning approaches are based on auto-encoded features optimized for specific datasets. For example, Breuer and Kimmel \cite{breuer2017deep} employ Ekman's facial action coding system (FACS) in order to boost the performances of their approach. On the other hand, deep learning results are clearly lower than handcrafted approaches in micro expressions recognition (Table \ref{synthese}). Furthermore, most of the time, in deep learning approaches, there is an augmentation of the initial data available (flip, rotate, Gaussian blur, etc.) in order to provide sufficient training data. Hence, comparison between deep learning approaches (using augmented data) and handcraft approaches (using only the original data) must be handled with care.

Reusing efficiently features and facial frameworks from macro expression to micro expression is not yet achieved with regard to the current state-of-the-art. The selected representative works employ the same underlying feature in micro and macro expression recognition, however they need to change the facial framework and the overall approach in order to maximize performances in both situations. Table \ref{synthese} shows that it is still difficult to find a common methodology to analyze both macro and micro expressions accurately. However, for both, dynamic approaches seems promising.

Starting from these observations, we propose an innovative motion descriptor called Local Motion Patterns (LMP) that keeps track of the real facial motion and filters the discontinuity of motion adapted for low and high amplitudes. Inspired by recent advances in the use of motion-based approaches for macro and micro expression recognition, we explore the use of magnitude and direction constraints in order to extract the relevant motion on the face. Considering the smoothing of motion in recent optical flow approach, simple optical flow combined with magnitude constraint is appropriate for reducing the noise induced by illumination changes and small head movements. In the next section, we propose to filter optical flow information based on consistent local motion propagation to keep only the pertinent motion
Then, in section \ref{sec:reconnaissance}, we explore the construction of a unified facial framework that generates discriminating features used to recognize effectively six macro expressions (anger, disgust, fear, happiness, sadness, surprise) and four micro expressions (disgust, happiness, repression, surprise).

\section{Local Motion Patterns}
\label{sec:filtrage}

The facial characteristics (skin smoothness, skin reflect and elasticity) induce inconsistencies when extracting motion information from the face. In our method, instead of explicitly computing the global motion field, the motion is computed in specific facial area, defined in relation with the facial action coding system in order to keep only the pertinent motion of the face. The pertinent motion is extracted from regions where the movement intensity reflects natural facial movements. We consider natural facial movement to be uniform and locally continuous over neighboring regions.

We propose a new feature named Local Motion Patterns (LMP) that retrieves the coherent motion around epicenter $\epsilon$(x,y) when considering natural motion propagation to neighboring regions.
Each region, called Local Motion Region (LMR), is characterized by a histogram of optical flow $H_{LMR_{x,y}}$, of B bins. There are two types of LMR involved: 
Central Motion Region (CMR), and 
Neighboring Motion Region (NMR).

\begin{figure}[!h]
\centering
\includegraphics[width=0.6\textwidth]{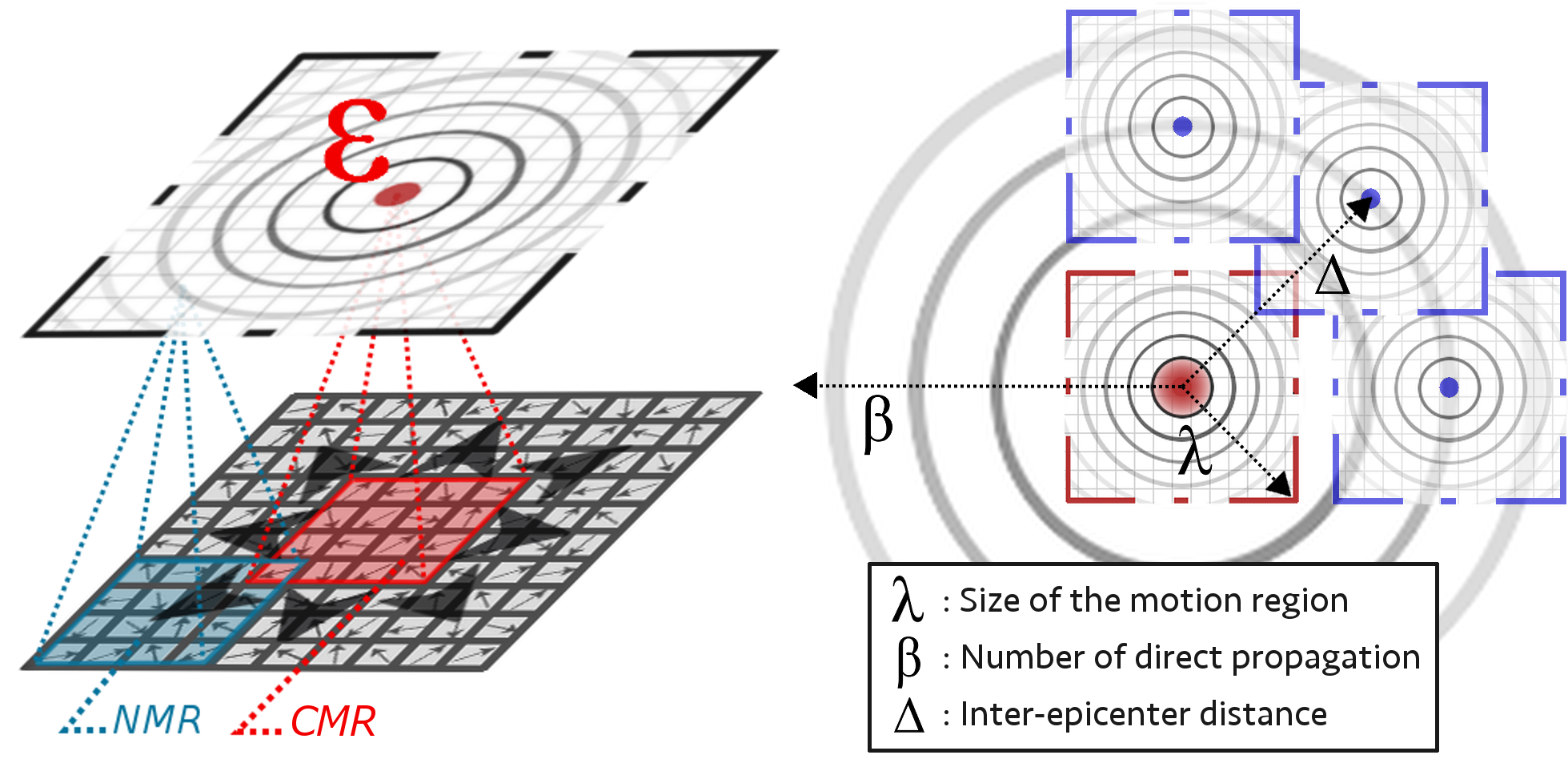}
\caption{Overview of local motion patterns extraction.}
\label{fig:LMP}
\end{figure}

LMP is illustrated in Figure \ref{fig:LMP}. Eight NMR are generated around the CMR. All these regions are at distance \textbf{$\Delta$} from the CMR. The bigger is the distance between two regions, the lesser is the coherence in the overlapping area. \textbf{$\lambda$} is the size of the area under investigation around the epicenter. Finally, $\mathbf{\beta}$ characterizes the number of direct propagations from the epicenter that are carried out by the motion propagation. The impact of these parameters on the quality of filtering process is detailed in section \ref{sec:evaluation}.

\subsection{Local coherency of central motion region}

In order to measure the consistency of the motion in terms of intensity and directions of LMP, we analyze the direction distribution in its CMR for several layers of magnitude. The motion of the face spreads progressively due to skin elasticity. We assume that there must be regular progression of magnitude in specific directions. 

Using filtering method of successive magnitude layers, the main motion directions tend to distinguish themselves from others, where the motion intensity is not continuous. Considering this observation, we propose a method to compute the direction in specific regions from different layers of magnitude. This technique brings out main directions that are much more difficult to analyze and reduces the motion noise.

The direction distribution of LMR is divided into $q$ histograms corresponding to different magnitude layers. 
The high intensity motion is more easily detected than low intensity motion 
. Each layer of magnitude extracted is defined as following:
\begin{equation}
\fontsize{8}{8}\selectfont 
MH_{LMR_{x,y}}(n,m) =  \{(bin_{i},mag_{i}) \in H_{LMR_{x,y}} \mid mag_{i} \in [n,m]\}.
\label{eq1}
\end{equation}

\noindent where $n$ and $m$ represent the magnitudes ranges and $i = 1,2,...,B$ is the index of bin.

Each $MH_{LMR_{x,y}}$ is normalized, and directions representing less than 10\%, are filtered out (set to zero). Then, magnitude layers are segmented into three parts $P_{1}\in [0\%,33\%]$, $P_{2}\in ]33\%,66\%]$ and $P_{3}\in ]66\%,100\%]$, represented by three cumulative histograms $ML_{LMR_{x,y}}(m1,m2)$ that are computed as follows:
\begin{equation}
\fontsize{8}{8}\selectfont 
\begin{split}
ML&_{LMR_{x,y}}(m1,m2) = \{(bin_{i}, 
card(\{(n,m) \mid \exists (bin_{i},mag_{i}) \in MH_{LMR_{x,y}}(n,m) \mid mag_{i} \in [m1,m2] \}) \text{ } ) \}.
\label{eq2}
\end{split}
\end{equation}

\noindent Finally, the directional and magnified histogram $DMH_{LMR_{x,y}}$ is determined by applying different weights to each part $\omega_{1}$, $\omega_{2}$ and $\omega_{3}$ of the corresponding bins, as follows:
\begin{equation}
\fontsize{8}{8}\selectfont 
\begin{split}
DMH_{LMR_{x,y}}
=  ML_{LMR_{x,y}}(m1,m2)*\omega_{1} + ML_{LMR_{x,y}}(m2,m3)*\omega_{2} 
+ ML_{LMR_{x,y}}(m3,m4)*\omega_{3}.
\label{eq3}
\end{split}
\end{equation}

\noindent We reinforce the local consistency of magnitude within each direction. We have applied 10-scale factor between layers ($\omega1 = 1$, $\omega2 = 10$ and $\omega3 = 100$). 
We assume that the higher is the intensity, the higher is the pertinence of motion.

\begin{figure}[!h]
\centering
\includegraphics[width=0.90\columnwidth]{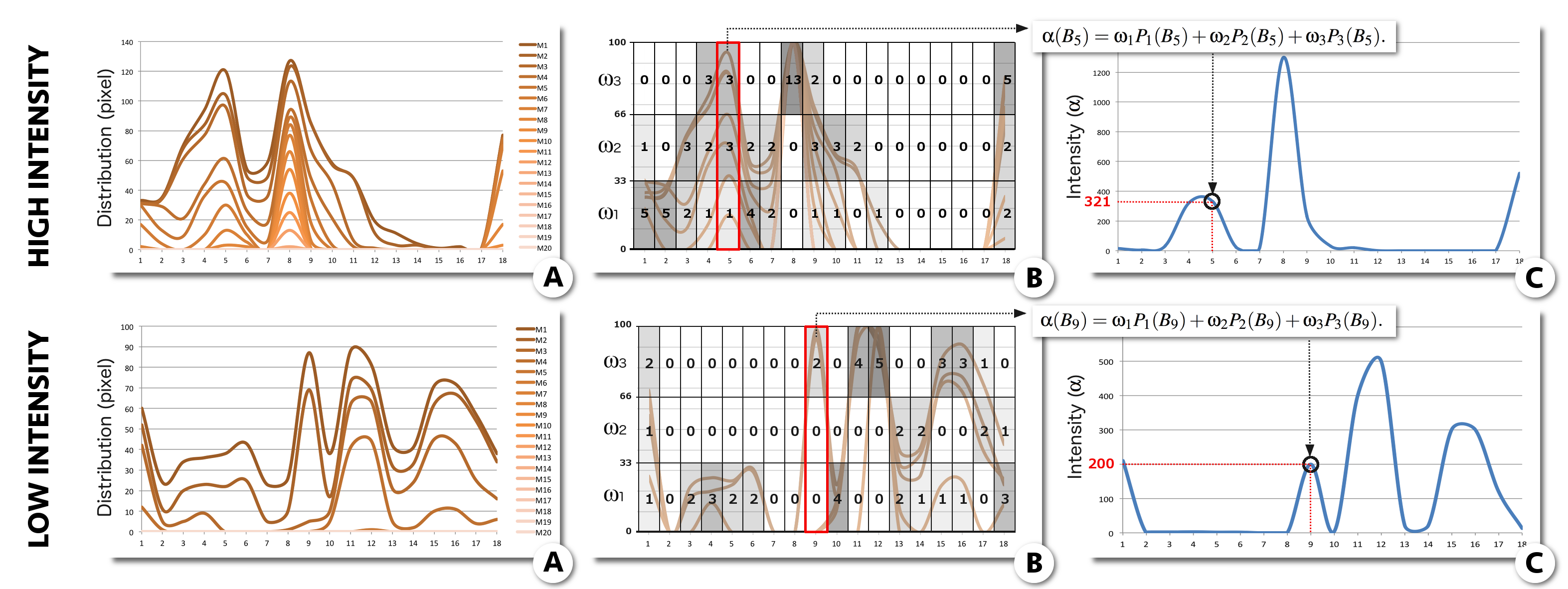}
\caption{The process of consistent local motion characterization in local motion region. (A) Magnitude histograms for different ranges, (B) Cumulative overlapping histograms, (C) Filtering motion.}
\label{fig:met}
\end{figure}

The motion filtering process is illustrated in Figure \ref{fig:met}
Figure \ref{fig:met}-A represents the histogram magnitude layers $MH_{LMR_{x,y}}$. Parameter $n$ is varying between 0 and 10, by 0.2 magnitude steps. The parameter $m$ is fixed to 10 in order to keep overlapping of magnitudes. The successive magnitude layers clearly distinguish the main direction. Next, the three magnitude layers $ML_{LMR_{x,y}}$ are represented in Figure \ref{fig:met}-B, where each $ML_{LMR_{x,y}}$ corresponds to a row, and the number in each cell represents the number of magnitude occurrences for each bin. Finally, directional and magnified histogram $DMH_{LMR_{x,y}}$ is illustrated in Figure \ref{fig:met}-C.

Before analyzing the neighborhood and confirming the coherency of LMR, at least one main direction in the distribution shall be obtained after applying a fixed threshold $\alpha$ to the directional and magnified histogram $(DMH_{LMR_{x,y}})$. The threshold value depends of the intensity importance in the filtering process. If no direction is found, LMR and LMP are locally incoherent. It means that the local intensity of motion does not exhibit expected progressive intensity for at least one direction.

Afterwards, it must be ensured that the main orientation directions into $DMH_{LMR_{x,y}}$ are consistent. In fact, the local distribution in LMR can be consistent in terms of intensity, and it is possible to have a large number of bins with high values. This step ensures that the local motions spread coherently in the local neighborhood.

In order to ensure consistent distribution in terms of orientation, the density of $k$ main directions is analyzed. Each main selected direction must satisfy several criteria. First criterion ensures that the main direction covers a limited number of bins (1 to $s$), where $s$ is the threshold for the number of bin spans accepted. Indeed, if we analyze a small region in a face, a coherent facial motion is rarely spreading over more than 60° and the variance of movement is progressive.
Otherwise, if one main direction is spreading over 60°, LMR stops analyzing the neighboring regions. Indeed, d
main directions spreading over 60° undermine the accurate identification of consistent motion by causing the propagation of false and misleading information. This criterion is defined by the following two equations. The first one characterizes the extend of main directions and the second filters out orientations spreading over $s$ consecutive bins:
\begin{equation}
\fontsize{8}{8}\selectfont 
\begin{split}
C(DMH_{LMR_{x,y}})
& = \{E=[a..b] \mid \forall_{i} \in [a..b] \mid DMH_{LMR_{x,y}}(i) > \alpha \\
& \text{ $\wedge$ } \nexists j \in \{a-1, b+1\} \mid DMH_{LMR_{x,y}}(j) > \alpha \}.
\end{split}
\label{eq4}
\end{equation}
\begin{equation}
\fontsize{8}{8}\selectfont 
C^{'}(DMH_{LMR_{x,y}}, s) = \{E \in C(DMH_{LMR_{x,y}}) \mid card(E) < s\}.
\label{eq5}
\end{equation}

\noindent where $[a..b]$ represents the limits that the standard deviation of directions must meet and $\alpha$ is the threshold value of the intensity. Then, for each selected direction, we keep only the directions spreading over at most $s$ consecutive bins.

In order to reinforce the fact that there is a gradual change in orientation, it is important that each main motion generates smooth transitions in terms of directions between neighbors. A maximum tolerance of $\Phi$ is supported 
as defined in the following:
\begin{equation}
\fontsize{8}{8}\selectfont 
\begin{split}
C^{''}(DMH_{LMR_{x,y}})
& = \{E=[a..b] \in C^{'}(DMH_{LMR_{x,y}},s) \mid \\
& \forall_{i,j} \in E, \norm{i-j} \leq 1 \mid \norm{DMH_{LMR_{x,y}}(i) - DMH_{LMR_{x,y}}(j)} < \text{ $\Phi$}\}.
\end{split}
\label{eq6}
\end{equation}

\noindent Finally, the filtered directional and magnified histogram $FDMH_{LMR_{x,y}}$ corresponds to $k$ main directions in $DMH_{LMR_{x,y}}$.
$FDMH_{LMR_{x,y}}$ is constructed as follows:
\begin{equation}
\fontsize{8}{8}\selectfont 
FDMH_{LMR_{x,y}} = \{ (b_{i}, m_{i}) \in DMH_{LMR_{x,y}} \mid \exists E=[a..b] \in C^{''}(DMH_{LMR_{x,y}}) \text{ $\wedge$ } b_{i} \in E \}.
\label{eq7}
\end{equation}

\noindent Despite CMR is considered coherent, LMP validation and computation have not yet been completed. Indeed, if we consider that natural facial movement is uniform during facial expressions, then the local facial motion should spread  over at least one neighboring region. 

\subsection{Neighborhood propagation of central motion region}

When LMP is locally coherent in CMR, the approach verifies the motion expansion on neighboring motion regions (NMR). In some cases, physical rules (e.g. skin elasticity) ensure that local motion spreads to neighboring regions until motion exhaustion. Motion is subject to changes that may affect direction and magnitude in any location. However, intensity of moving facial region tends to remain constant during facial expression. Therefore, a pertinent motion observed and computed in CMR appears, eventually with lower or upper intensity, in at least one neighboring region.

Before analyzing the motion propagation, the local coherency of each NMR is analyzed with the same method discussed above for CMR. As an outcome of the process, each locally consistent NMR is characterized by LMR. As for CMR, it must be ensured that the local distribution is consistent in terms of intensity and orientation. However, before considering the orientation consistency, it is important to check that the local distribution is similar to some extent with the previous adjacent neighbor. 
Bhattacharyya coefficient is used to measure the overlap between two neighboring LMR and it is computed as follows:
\begin{equation}
\fontsize{8}{8}\selectfont 
C^{'''}(FDMH_{LMR_{x,y}},FDMH_{LMR_{x,y}}^{'}) = \sum_{i=1}^B \sqrt[]{FDMH_{LMR_{x,y}}(i)FDMH_{LMR_{x,y}}^{'}(i)}.
\label{eq8}
\end{equation}

\noindent where, $FDMH_{LMR_{x,y}}$ and $FDMH_{LMR_{x,y}}^{'}$ are the local distributions and $B$ is the number of bins. LMR is considered consistent with his neighbor, if the coefficient exceeds the fixed threshold $\rho$.

\begin{figure}[!h]
\centering
\includegraphics[width=0.8\textwidth]{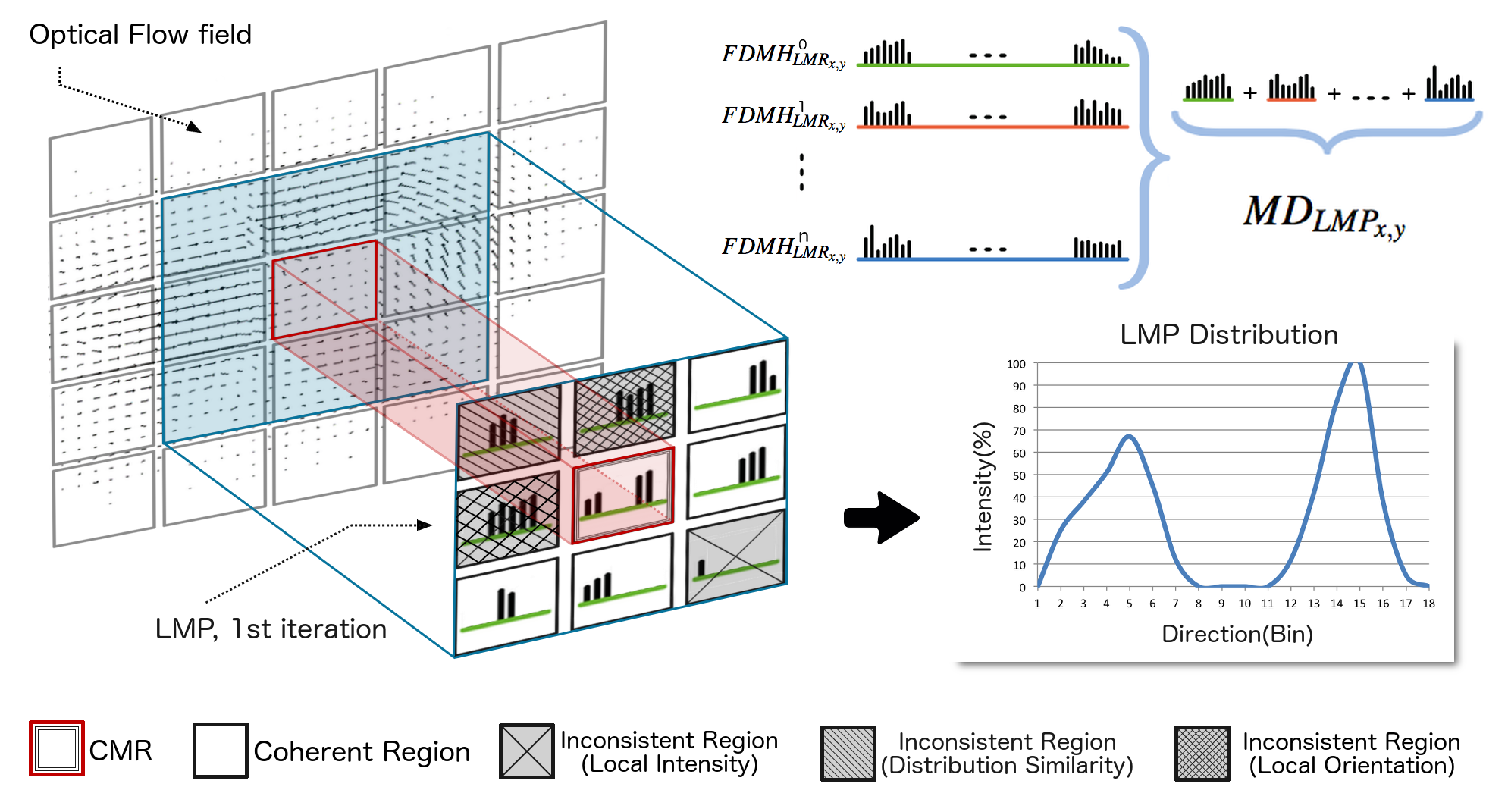}
\caption{LMP distribution, computed from the propagation in neighborhood of central motion region.}
\label{fig:propagation}
\end{figure}

The motion propagation into LMP after one iteration is given in Figure \ref{fig:propagation}. NMR are represented in gray, if the motion is inconsistent. Three situations, where NMR are not coherent, can be distinguished: a) The motion in NMR is locally inconsistent in terms of intensity; b) The distribution similarity between two regions is considered inconsistent and c) The motion in NMR is locally inconsistent in terms of orientation.

As long as at least one newly created NMR is inter-region coherent with its neighbor, recursively, for each subsequent NMR, the motion analysis is conducted in the same way. The recursive process ends when the number of propagations $\beta$ is reached.

Finally, each distribution ($FDMH_{LMR_{x,y}}$) corresponding to NMR that have direct or indirect connections to original CMR contributes to the LMP distribution. If the motion propagation between all NMR is inconsistent, the motion propagation is no more explored. This means that there are no more pertinent motions to collect into LMP. The motion distribution of LMP is extracted by applying the following formula:
\begin{equation}
\fontsize{8}{8}\selectfont 
MD_{LMP_{x,y}} = \{ \sum_{i=0}^{n} FDMH_{LMR_{x,y}} \mid FDMH_{LMR_{x,y}} \in LMP_{x,y} \}.
\label{eq9}
\end{equation}

\noindent where $n$ is the number of consistent regions (CMR and all consistent NMR). The maximum value of $n$ is computed by the following formula:
\begin{equation}
\fontsize{8}{8}\selectfont 
Max(n) = 1 + c * \frac{ \beta ( \beta + 1)}{2} \text{ if } \beta \geq 1  \text{ , } 1 \text{ else}.
\label{eq:beta}
\end{equation}

\noindent where $\beta$ is the number of iterations and $c$ corresponds to the pixel connectivity (here $c=8$). $MD_{LMP_{x,y}}$ is characterized by histogram over $B$ bins, which contains, for each bin the sum of intensities collected from coherent NMR and CMR. Then we are able to extract the coherent motion from a specific location on the face.

In summary, the proposed local motion patterns feature collects pertinent motion and filters the noisy motion based on three criteria: the convergence of motion intensity in the same direction, the local coherency of direction distribution and the way the motion spreads. Each criterion can be configurable independently of the others, which makes it fully adaptable to many usages as action recognition, facial expression recognition, tracking and other. To prove the effectiveness of our LMP, we analyze in the next section, the use of LMP for micro and macro facial expression analysis.

\section{Expression recognition}
\label{sec:reconnaissance}

The choice of the facial recognition framework impacts greatly the performances. Various epicenters can be considered for coherent motion extraction. So, we study the impact of epicenters on the perceived motion while applying LMP. We show that the intensity of expression (macro or micro) plays a key role in locating LMP epicenter and, in the same time, it impacts the way the consistent motion of the face is encoded. Then, we explore the integration of the coherent optical flow into facial model formulation, and discuss several strategies for considering discriminant local regions of the face.

\subsection{Impact of LMP location}

\begin{figure}[!h]
\centering
\includegraphics[width=0.70\columnwidth]{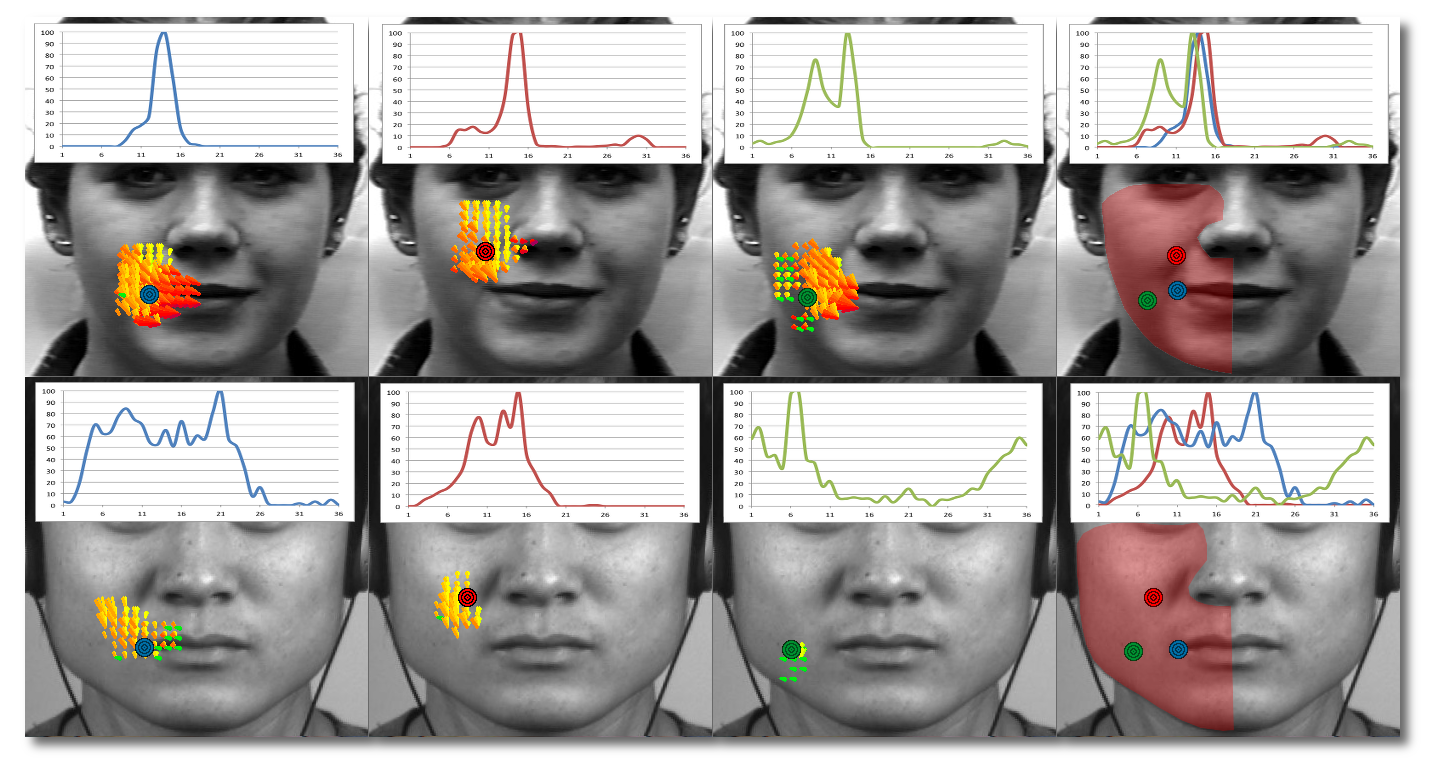}
\caption{Consistent motions from happiness sequence computed from different locations in the same region.}
\label{fig:mot}
\end{figure}

In macro expression, motion propagation covers large facial area. If one CMR (Central Motion Region) is randomly placed in this area, then the motion consistency is most of the time observed. However, in micro expression, the motion propagation covers restricted facial area. Motions are less intense, so motion propagation is discontinued.
Figure \ref{fig:mot} shows local motion distribution extracted in various points around left lip corner (blue, red and green dots). The original flow field and the local motion distribution extracted from a happiness sequence around the different locations are shown in the first three columns. The fourth column shows the overlap of previous ones.

For macro expression (first line), location of each LMR is different, the distributions present large overlaps (column 4). For micro expression (second line), the distributions corresponding to the three columns are different. The experimentation can be reproduced in other facial regions with similar outcomes with regard to micro versus macro expressions. It is hence important to determine best discriminant facial regions for encoding coherent motion in the context of generic expression recognition process.


\subsection{Best discriminant facial region}

Macro and micro expression motions are very different in terms of intensity and propagation. It is therefore important to detect pertinent motions that generate features able to discriminate effectively some of the most common macro expressions (happiness, sadness, fear, disgust, surprise and anger) and micro expressions (happiness, disgust, surprise, repression). In order to identify optimal LMP epicenters locations, we have considered samples for CK+ and CASME2 datasets.

\begin{figure}[!h]
\centering
\includegraphics[width=0.8\columnwidth]{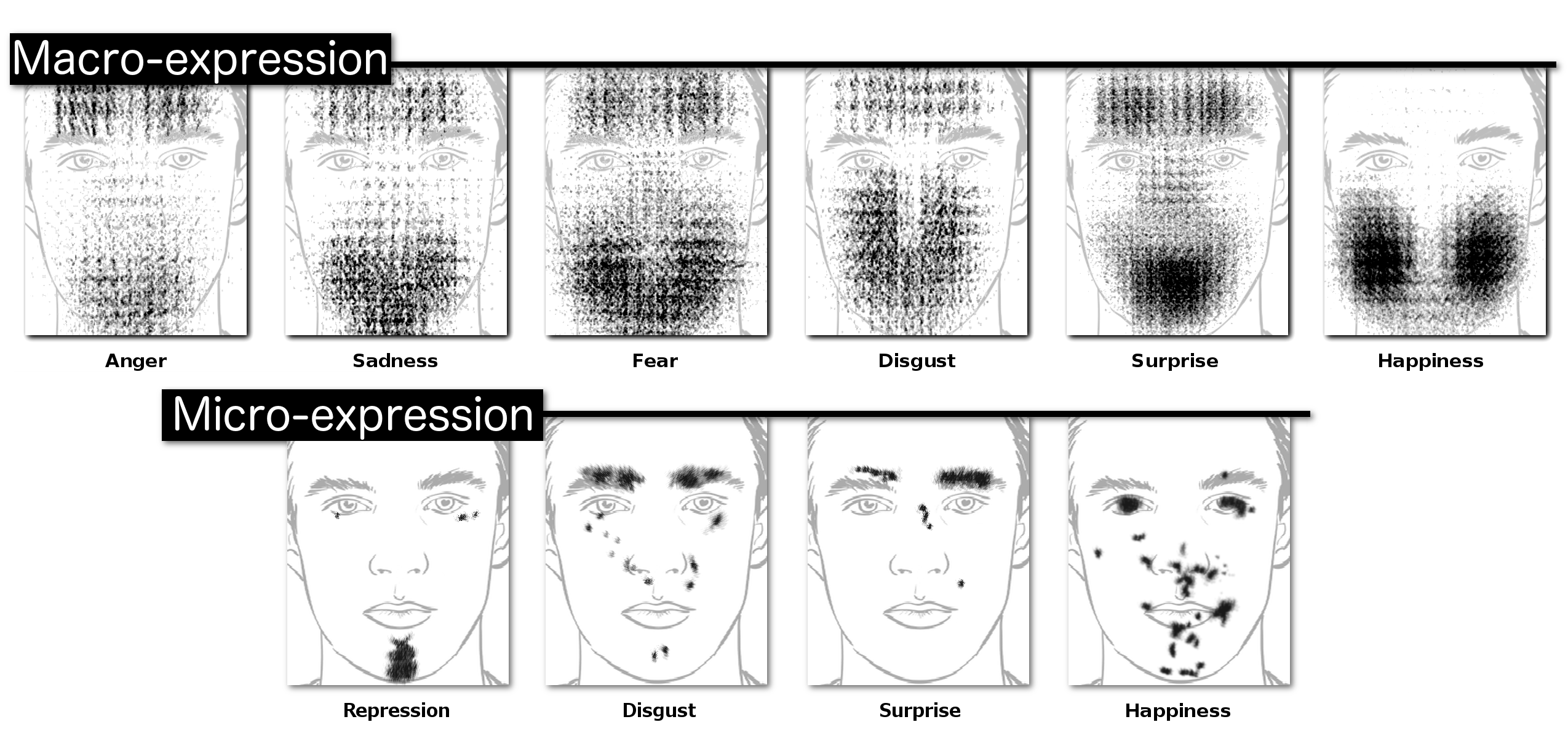}
\caption{Pertinent motions elicitating six macro expressions from CK+ dataset (line 1) and four micro expressions from CASME II dataset (line 2).}
\label{fig:map1}
\end{figure}

To identify the locations within the face where motion often occurs, we first align frames based on eyes location, and we compute the optical flow of each frame of the sequence. Then, each frame is segmented in $20x30$ blocks. This step eliminates in-plane head rotation and addresses individual differences in face shape. LMP is associated to each block, and CMR are located in their respective centers. Then, the consistent motion vector is computed in each LMP. Next, each relevant optical flow extracted from each frame is merged into a single binary motion mask. The consistent motion mask as well as motion information are extracted from  video sequences of the same expression class. Finally, each consistent motion mask is normalized and merged to form a heat map of motion for the underlying expression. The six consistent motion masks for the basic macro expressions are illustrated in the first line of Figure \ref{fig:map1}. They are computed from the sequences available in CK+ dataset.

The extracted mask indicates that pertinent motions are located below the eyes, in the forehead, around the nose and mouth, as illustrated in Figure \ref{fig:map1}. Some facial motions are located in the same place during elicitation for several expressions, but they are distinguishable by their intensity, direction and density. For example, anger and sadness motion masks are similar as the main motion appears around the mouth and the eyebrows. However, when a person is angry, motion is convergent (e.g mouth upwards and eyebrows downwards), and  motion is divergent when a person is sad.

The same search strategy for finding the best discriminant regions for macro expressions in CK+ dataset was used in CASME II dataset for micro expressions (happiness, disgust, surprise and repression). As illustrated in the second line of Figure \ref{fig:map1}, the pertinent motions are located near the eyebrows and the lips corner. If we compare with macro expression motion maps, propagation distances are highly reduced.

At this stage, the main facial regions of motion are accurately identified. We now construct a vector that expresses the relationships between facial region of motion and expressions. We use the facial landmarks to define regions that increase deformation robustness during expression. Similarly to Jiang et al. \cite{jiang2014decision}, the landmarks are used to define a mesh over the whole face, and a feature vector can be extracted from the regions enclosed by the mesh. 
Landmarks and geometrical features of the face are used to compute the set of points that defines a mesh over the whole face (forehead, cheek). Finally, the best discriminant landmarks are selected corresponding to active face regions, and specific points are computed in order to set out the mesh boundaries.

\begin{figure}[!h]
\centering
\includegraphics[width=0.5\columnwidth]{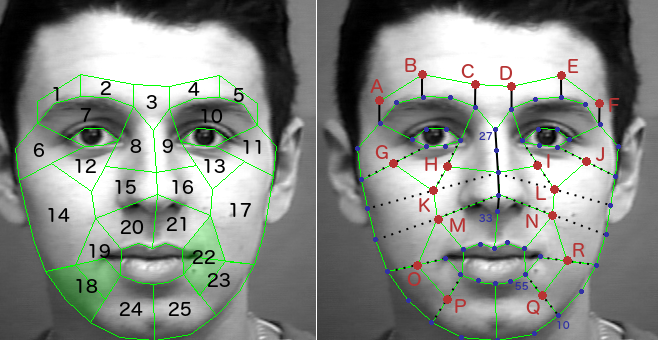}
\caption{Facial partition in interest regions.}
\label{fig:mask}
\end{figure}

The partitioning into facial regions of interest (ROIs) is illustrated in Figure \ref{fig:mask}. The partitioning is based on the facial motion observed in the consistency maps constructed from both macro and micro expressions. The locations of these ROIs are uniquely determined by landmarks points for both micro and macro expressions. For example, the location of feature point $P_{Q}$ is the average of two feature points, $P_{10}$ and $P_{55}$. The distance between eyebrows and forehead feature points ($P_{A}$,$P_{B}$,...,$P_{F}$) corresponds to the size of the nose $Distance_{P_{27},P_{33}}/4$. This allows maintaining the same distance for optimal adaptation to the size of the face. Note that, in order to deal precisely with the lip corners motion, regions 19 and 22 overlap regions 18 and 23, respectively.

\subsection{Facial motion descriptor}

The facial motion mask is computed from the 25 ROIs presented above. In each frame $f_{t}$, we consider the filtered distribution motion inside each ROI $R_{t}^{k}$, where $t$ is the frame index and $k = 1,2,...,25$ is the ROI index. Inside each $R_{t}^{k}$, LMP is applied and $MD_{LMP_{x,y}}$ is computed as defined in equation \ref{eq9}. $R_{t}^{k}$ motion distributions are summed into $\eta^{k}$, which corresponds to local facial motion in region $k$ for the entire sequence.
\begin{equation}
\fontsize{8}{8}\selectfont 
\eta^{k} = \sum_{t=1}^{time} R_{t}^{k}.
\label{eq10}
\end{equation}

\begin{figure}[!h]
\centering
\includegraphics[width=0.80\columnwidth]{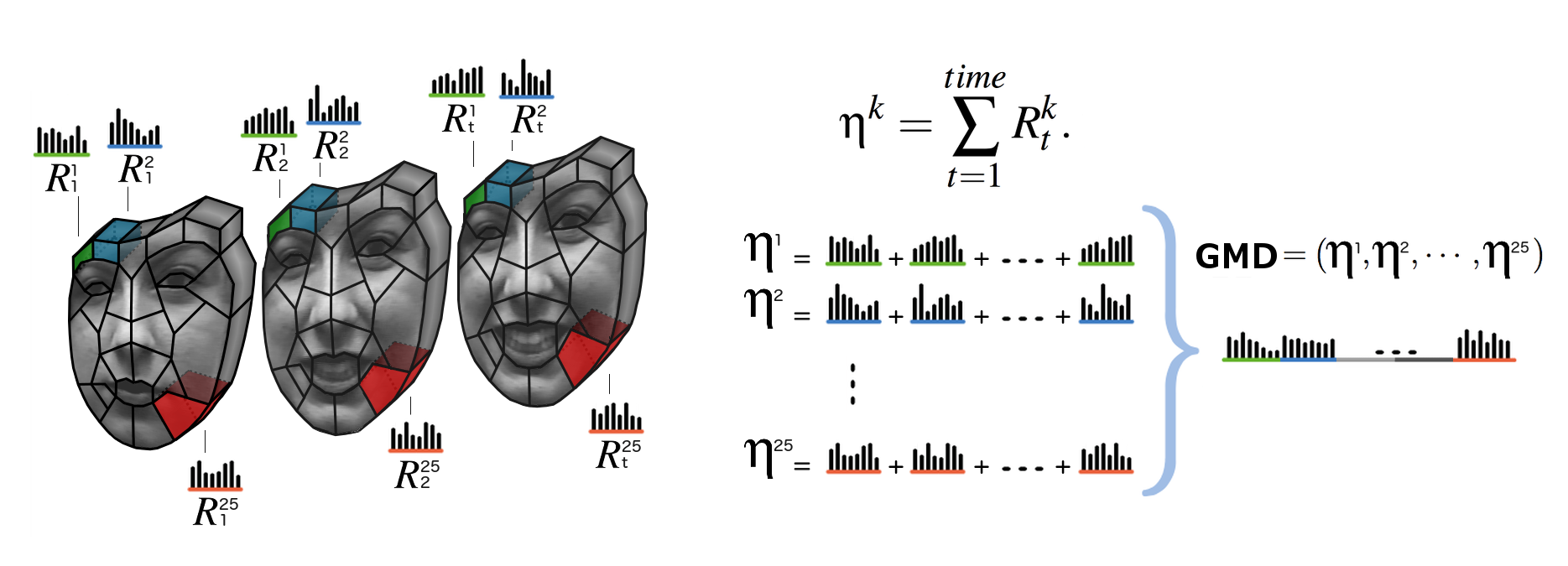}
\caption{Building the feature vector from the facial motion mask.}
\label{fig:vector}
\end{figure}

\noindent Finally, histograms $\eta^{k}$ are concatenated into one-row vector $GMD$, which is considered as the feature vector for the macro and micro expression $GMD = (\eta^{1},\eta^{2},...,\eta^{25})$. The feature vector size is equal to the number of ROI multiplied by the number of bins. An example is illustrated in Figure \ref{fig:vector}, where all motion distributions $MD_{LMP_{x,y}}$ corresponding to $R_{t}^{1}$, $R_{t}^{2}$ ... $R_{i}^{25}$ with $t\in[1,time]$ are summed up in $\eta^{1}$, $\eta^{2}$ ... $\eta^{25}$ respectively. $\eta^{1}$, $\eta^{2}$ ... $\eta^{25}$ are then concatenated, and define the global motion distribution $GMD$.

\subsection{Facial Recognition framework}

The framework, presented in Figure \ref{fig:schemeT1}, is suitable to micro and macro expressions and is composed of  feature extraction and recognition tasks.
Feature extraction consists in pre-processing, that detects facial landmarks. 
The Farnebäck algorithm \cite{farneback2003two} is used to compute fast dense optical flow. It ensures that motion is not affected by smoothing and the computation time is low. ROI are defined and local motion patterns features are associated to each ROI. Then, the consistent motion vector of each local motion pattern is computed. Motion filtering and analysis, illustrated in Figure \ref{fig:schemeT1}, are the main contributions of our method. Next, relevant motion in each facial region is cumulated over time. Each facial region is represented by a histogram based on the orientation and the intensity of motion. Classification is performed on features extracted from videos.


\section{Evaluation}
\label{sec:evaluation}

We highlight the performance of our proposed method on widely used datasets for micro expression recognition, namely CASME II \cite{yan2014casme} and SMIC \cite{li2013spontaneous}, and widely used datasets for macro expression recognition, namely CK+ \cite{lucey2010extended}, Oulu-CASIA \cite{zhao2011facial}, and MMI \cite{pantic2005web}. Experiments and comparisons on these datasets cover aspects of in-the-wild recognition, such as head movement, illumination, visible and infrared contexts.

After introducing the datasets, we discuss the impact of our method parameters (size of the motion region, number of direct propagations, ...) on performances. We show that only few parameters need to be tuned in order to accommodate micro and macro expressions. Then, we compare our performances with major state-of-the-art approaches. We use LIBSVM \cite{chang2011libsvm} with RBF kernel and 10 fold cross-validation protocol for macro expression and leave-one-subject-out for micro expression.

\subsection{Datasets}

\paragraph{CASME II} (micro expression dataset) contains 247 spontaneous micro expressions from 26 subjects, categorized into five classes: happiness (33 samples), disgust (60 samples), surprise (25 samples), repression (27 samples) and others (102 samples). The micro expressions are recorded at 200 fps in well-controlled laboratory environment.

\paragraph{SMIC} (micro expression dataset) is divided into three sets
: (i) HS dataset is recorded by high-speed camera at 100 fps and includes 164 sequences from 16 subjects, (ii) VIS dataset is recorded by standard color camera at 25 fps; and (iii) NIR dataset is recorded by near infrared camera at 25 fps. The high-speed (HS) camera was used to capture and record the whole data, while VIS and NIR cameras were only used for recording the last eight subjects (77 sequences). The three datasets include micro expression clips of videos from onset to offset. Each sequence is segmented from the original videos and labeled with three emotion classes: positive, surprise and negative.

\paragraph{CK+} (macro expression dataset) contains 593 acted facial expression sequences from 123 participants, with seven basic expressions (anger, contempt, disgust, fear, happiness, sadness, and surprise). In this dataset, the expression starts at neutral state and finishes at apex state. Expression recognition is completed in excellent conditions, because the deformations induced by ambient noise, facial alignment and intra-face occlusions are not significant with regard to the deformations directly related to the expression. 

\paragraph{Oulu-CASIA} (macro expression dataset) includes 480 sequences of 80 subjects taken under three different lighting conditions: strong, weak and dark illuminations. They are labeled with one of the six basic emotion labels (happiness, sadness, anger, disgust, surprise, and fear). Each sequence begins with neutral facial expression and ends with apex. In-the-wild conditions, varying lighting conditions influence the recognition process. Expressions are simultaneously captured in visible light and near infrared.

\paragraph{MMI} (macro expression dataset) includes 213 sequences from 30 subjects. The subjects were instructed to perform six expressions (happiness, sadness, anger, disgust, surprise, and fear). Compared with CK+ and Oulu-CASIA, due to strong head pose variations of subjects, MMI is more challenging. Subjects are free of their head and expressions show similarities with in-the-wild settings.

\subsection{Framework parameters}
\label{sec:parameters}

In this section, we review the parameters that are involved in the motion filter process. We propose a methodology for finding the best parameters depending of the analyzed datasets (CK+ for macro expression and CASME II for micro expressions).

\begin{figure}[!h]
\centering
\includegraphics[width=\textwidth]{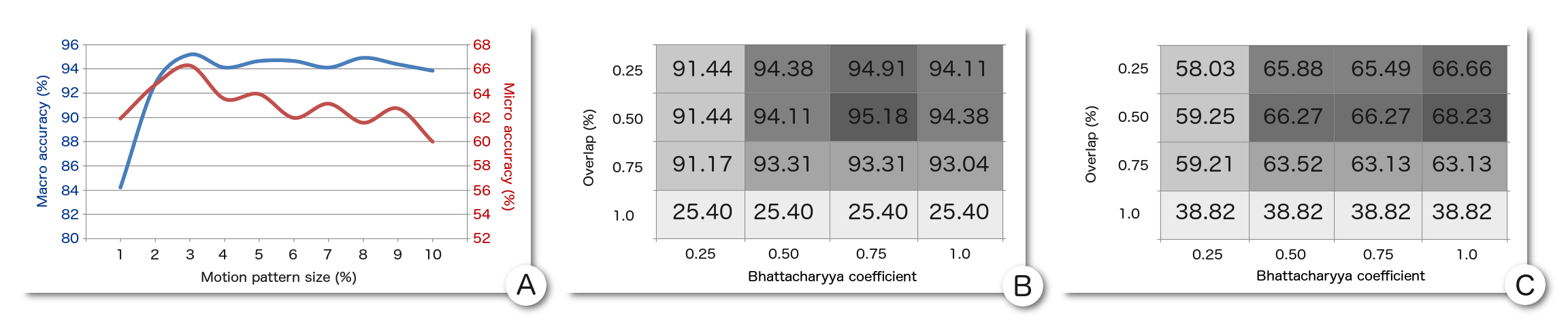}
\caption{Recognition rates for different $\lambda$ (A), and the relation between the overlap $\Delta$ and Bhattacharyya coefficient $\rho$ of macro (B) and micro expressions (C).}
\label{fig:schemeF1}
\end{figure}

The first parameter defines LMP region size as a percentage ($\lambda$) of the whole face. This guarantees that LMP regions share common characteristics regardless of variations in terms of face size within a video sequence. As illustrated in Figure \ref{fig:schemeF1}-A, the ideal size of $\lambda$ is around 3 percent of the face size, for both macro and micro expressions. LMP region is lower enough (3\%) in order to support correctly motion propagation. A large region is characterized by a sparse motion distribution, which does not distinguish the main direction and can reduce the quality of motion filtering over time.

The second parameter is the spatial overlap $\Delta$ between the neighboring LMP regions. To ensure that the motion spreads progressively (intensity and direction), it is important that neighboring regions overlap. As illustrated in Figures \ref{fig:schemeF1}-B and \ref{fig:schemeF1}-C, an overlap of 50\% gives the best performances for macro and micro expressions. According to that, it is possible to define the most adapted coefficient to distinguish the Bhattacharyya distance of distribution $\rho$ from two overlapped regions.

\begin{figure}[!h]
\centering
\includegraphics[width=0.8\textwidth]{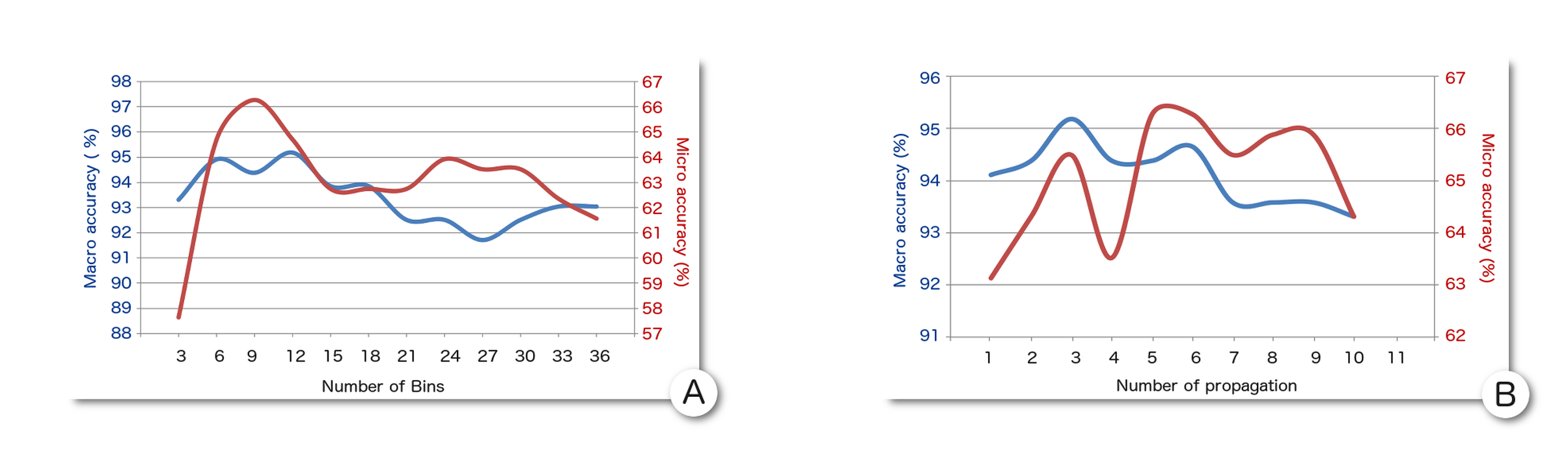}
\caption{Recognition rates from different distribution sizes (A) and different number of propagations (B).}
\label{fig:schemeF2}
\end{figure}

Other parameters need to be taken into account such as the number of bins $B$ and the number of iterations $\beta$, which is the number of propagations from the current region to the overlapped neighborhood. When $\lambda$ is small, it is more interesting to take a small number of bins, which distinguishes more easily the main direction. As shown in Figure \ref{fig:schemeF2}-A, the best performances are obtained using histograms with 9 or 12 bins. Concerning the propagation in Figure \ref{fig:schemeF2}-B, increasing the number of iterations improves the performances to some extent for micro expressions. Indeed, LMP distribution, that considers several neighbor regions, allows better discrimination of similar facial expressions. When the number of iterations ($\beta$) is beyond a certain level, then LMP becomes noisy. This is especially true in presence of macro expressions due to the quick spread of the motion in different directions.

\begin{figure}[!h]
\centering
\includegraphics[width=\textwidth]{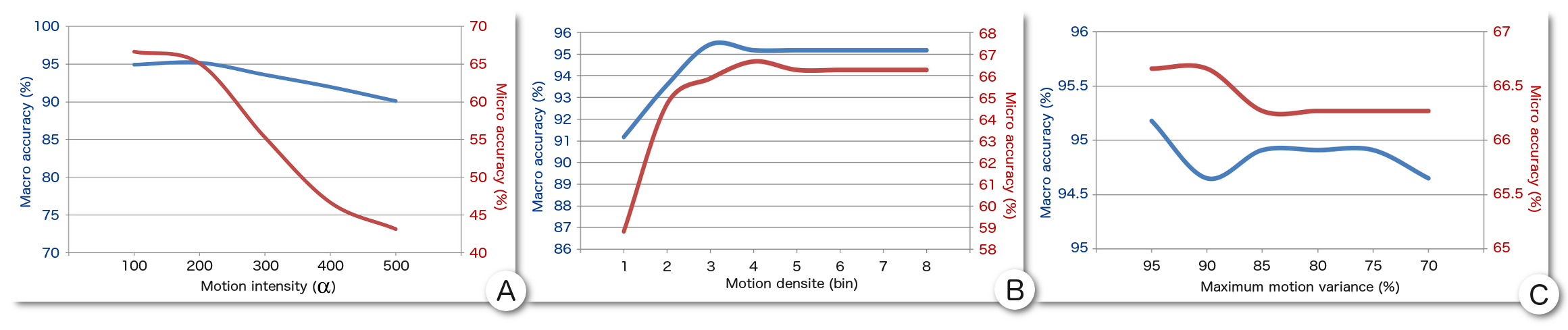}
\caption{Recognition rates from different motion intensity (A), motion density (B), motion variation (C).}
\label{fig:schemeF3}
\end{figure}

Finally, three parameters directly linked to the desired quality of the filtering process are considered. Depending on the importance according to the intensity of the direction, we calculate threshold describing coherent motion, on the basis of the motion intensity $E$, the variance between neighbor regions $M$ and the maximum tolerance accepted between two bins $V$. Concerning the motion intensity, illustrated in Figure \ref{fig:schemeF3}-A, the bigger the motion intensity threshold is, the lower the performance is. This is particularly true for micro expressions because the direction magnified histogram value does not exceed $\alpha$ value which is higher than 200 (computed with Equation \ref{eq3}). As shown in Figure \ref{fig:schemeF3}-B, filtering applied on motion density improves the performance rate. Finally, the motion variation between bins does not have big impact, compared to other parameters as illustrated in Figure \ref{fig:schemeF3}-C.

On the basis of framework parameters (e.g. size of the motion region, number of direct propagations, inter-epicenter distance), it is important to note that LMP can be easily adapted, and parameters have a moderated impact of the motion filtering process for micro and macro expression recognition.


\subsection{Micro expression}

In this section, we show the experiment results on CASME II and SMIC micro expression datasets, followed by discussion and analysis of the results.

\paragraph{Experiments on CASME II} Table \ref{state3} shows a comparison between major state-of-the-art micro expressions approaches.
In our method, the optical flow is calculated from two consecutive frames. The geometric information is not considered here, as landmarks locations are mostly stable throughout the sequence. We have selected only the activation sequence (e.g. onset to apex) for each clip.

\begin{table}[!h]
\renewcommand{\arraystretch}{1.3}
\caption{Performance comparison on CASME II dataset \textit{(* Data augmentation / Deep Learning})}.
\label{state3}
\centering
\fontsize{7}{5}\selectfont
\begin{tabular}{lcccc}
\hline \multicolumn{1}{c}{Method} & Interpolat. & Magnifi. & Classes & Acc(\%) \\
\hline Baseline \cite{yan2014casme}  & \ding{55} & \ding{55} & 5 & 63.41\% \\
LBP-SIP \cite{wang2014lbp} & \ding{55} & \ding{55} & 5 & 67.21\% \\
LSDF \cite{wang2014micro} & \ding{51} & \ding{55} & 5 & 65.44\% \\
STCLQP \cite{huang2016spontaneous} & \ding{51} & \ding{55} & 5 & 58.39\% \\
STLBP-IIP \cite{huang2016spontaneous2} & \ding{55} & \ding{55} & 5 & 62.75\% \\
DiSTLBP-IPP \cite{huang2016spontaneous2} & \ding{55} & \ding{55} & 5 & 64.78\% \\
HIGO \cite{li2015reading} & \ding{51} & \ding{51} & 5 & 67.21\% \\
MDMO \cite{liu2015main} & \ding{55} & \ding{55} & 4 & 67.37\% \\
CNN + LSTM \cite{kim2017multi} & \ding{55} & \ding{55} & 5 & * 60.98\% \\
CNN + AUs + LSTM \cite{breuer2017deep} & \ding{55} & \ding{55} & 5 & * 59.47\% \\
\textbf{Local Motion Pattern (LMP)} & \ding{55} & \ding{55} & \textbf{5} & \underline{\textbf{70.20\%}} \\
\hline
\end{tabular}
\end{table}

In view of the results obtained in Table \ref{state3}, our method outperforms the other state-of-the-art methods, including handcrafted and deep methods, in all cases. Looking closely, some authors summarize videos in fewer frames \cite{wang2014micro,li2015reading}. Indeed, the time lapse between two frames in CASME II is very small (recorded with a high-speed camera (at 200 fps) and combined with the low expression intensity, it makes the distinction between the noise and the true facial motion very difficult. In \cite{li2015reading} a magnification process, which consists of interpolating the frequency, in order to intensify the facial motion is used. These techniques perform well in presence of low intensity motion, but produce severe facial deformations in presence of high intensity motions or head pose variations.
Even-though deep learning methods \cite{kim2017multi,breuer2017deep} employ data augmentation, their performances are clearly lower that then those of handcrafted methods. This is mainly due to the characteristics of the data of CASME II dataset, in which there does not exist a large intra-class variation between different subjects.
The results obtained in the CASME II dataset show good performances for micro expressions recognition where no illumination changes appear. In the next paragraph, we evaluate our method on micro expression in presence of various illumination settings.

\paragraph{Experiments on SMIC} Table \ref{stateSMIC} compares the performance of the proposed method with major state-of-the-art approaches on SMIC dataset under three different acquisition conditions: sequences recorded by high-speed camera at 100 fps (HS), sequences recorded by normal color camera at 25 fps (VIS) and sequences recorded by a near infrared camera both at 25 fps (NIR).

\begin{table}[!h]
\renewcommand{\arraystretch}{1.3}
\caption{Performance comparison on SMIC dataset \textit{(* Data augmentation / Deep learning)}.}
\label{stateSMIC}
\centering
\fontsize{7}{5}\selectfont
\begin{tabular}{lcccc}
\hline \multicolumn{1}{c}{Method} & Magnifi. & SMIC-HS & SMIC-VIS & SMIC-NIR \\
\hline
LBP-TOP \cite{li2013spontaneous} & \ding{55} & 48.78\% & 52.11\% & 38.03\% \\
Selective Deep features (CNN) \cite{patel2016selective} & \ding{55} & * 53.60\% & * 56.30\% & N/A \\
Facial Dynamics Map \cite{xu2017microexpression} & \ding{55} & 54.88\% & 59.15\% & 57.75\% \\
HIGO \cite{li2015reading} & \ding{55} & 65.24\% & 76.06\% & 59.15\% \\
HIGO \cite{li2015reading} & \ding{51} & \underline{68.29\%} & 81.69\% & 67.61\% \\
\textbf{Local Motion Patterns (LMP)} & \ding{55} & \textbf{67.68\%} & \underline{\textbf{86.11\%}} & \underline{\textbf{80.56\%}} \\
\hline
\end{tabular}
\end{table}

Our method outperforms the state-of-the-art methods, including handcrafted and deep learning methods, in all cases, with similar performance in SMIC-HS subset. Li et al. \cite{li2015reading} show that artificially amplifying the motion tends to improve the results for micro expression recognition. However, interpolating the video frequency cannot be appropriately generalized on macro expression.
The results obtained on SMIC dataset show good performances for micro expressions recognition with regard to near infrared and natural illumination conditions. Our method based on optical flow seems to be better adapted in near infrared condition compared to other dynamic methods.


\subsection{Macro expression}

In this section, we study the performance of our method to recognize macro expressions on CK+, Oulu-CASIA and MMI datasets dealing respectively with variations in temporal activation sequences, illumination variations and head movements.

\paragraph{Experiments on CK+} Table \ref{stateCK} compares the performance of the proposed method with major state-of-the-art approaches on CK+ dataset. Despite CK+ dataset is the most widely used for evaluating performance, it is especially difficult to compare with other approaches because they generally do not use the same experimental protocol (e.g. number of subjects, classes). We use the two most representative subsets to evaluate the performances of our method in CK+ dataset. The first subset contains 327 sequences that include seven expressions as follows: anger (45), sadness (28), happiness (69), surprise (83), fear (25), disgust (59) and contempt (18). The second subset contains 374 sequences based only on the six universal facial expressions. This second subset contains the following number of samples per expressions: anger (37), sadness (65), happiness (95), surprise (83), fear (53) and disgust (41).

\begin{table}[!h]
\renewcommand{\arraystretch}{1.3}
\caption{Performance comparison on CK+ dataset \textit{(* Data augmentation / Deep learning)}.}
\label{stateCK}
\centering
\fontsize{7}{5}\selectfont
\begin{tabular}{lcccc}
\hline
\multicolumn{1}{c}{Method} & Measure & Seq. & Classes & Acc(\%) \\
\hline
Spatial weight mask \cite{liao2013learning} & LOSO & 442 & 6 & 92,50\% \\
Optical flow and three MLPs \cite{su2007simple} & train/test & 415 & 5 & 93,27\% \\
\textbf{Local Motion Patterns (LMP)} & \textbf{10-fold} & \textbf{327} & \textbf{7} & \textbf{96.94\%} \\
\textbf{Local Motion Patterns (LMP)} & \textbf{10-fold} & \textbf{374} & \textbf{6} & \textbf{96.26\%} \\
LBP-TOP + Gabor \cite{zhao2017facial} & LOO & 309 & 6 & 95.80\% \\
Dis-ExpLet \cite{liu2016learning} & 10-fold & 327 & 7 & 95.10\% \\
LBP-TOP \cite{zhao2007dynamic} + manual alignment & 10-fold & 374 & 6 & 96.26\% \\
Spatio-temporal RBM-based model \cite{elaiwat2016spatio} & 10-fold & 327 & 7 & 95.66\% \\
CNN + AUs + LTSM \cite{breuer2017deep} & LOO & \textit{augmented} & 7 & * \underline{98.62\%} \\
CNN + Spatial features \cite{lopes2017facial} & 8-fold & \textit{augmented} & 7 & * 96.76\%  \\
CNN + Spatial features \cite{lopes2017facial} & 8-fold & 327 & 7 & * 86.67\%  \\
PHRNN-MCSNN \cite{zhang2017facial} & 10-fold & \textit{augmented} & 7 & * \underline{98.50\%}  \\
DTAGN (joint) \cite{jung2015joint} & 10-fold & \textit{augmented} & 7 & * 97.25\%  \\
\textbf{LMP + Geometric features} & \textbf{10-fold} & \textbf{327} & \textbf{7} & \textbf{97.25\%} \\
\textbf{LMP + Geometric features} & \textbf{10-fold} & \textbf{374} & \textbf{6} & \textbf{96.79\%} \\
\hline
\end{tabular}
\end{table}

Compared to optical flow approaches \cite{liao2013learning,su2007simple} and handcrafted approaches \cite{zhao2017facial,fan2015spatial,fan2017dynamic,liu2016learning,elaiwat2016spatio}, our method based only on optical flow obtains competitive results (96.94\%). Despite the noise contained in the original optical flows, the variation in sequence length and expression activation patterns, the joint analysis of magnitudes and orientations keeps only the pertinent motion.

Inspired by improvements obtained by hybrid approaches, we combine motion features with geometric features by exploiting the shape of facial ROIs for the apex frame. Combination of geometric and LMP features improves slightly the results (97.25\%).

Results of recent deep learning approaches \cite{lopes2017facial,breuer2017deep,zhang2017facial,jung2015joint} obtained on CK+ are comparable with the best results that we obtained using a handcrafted approach. However, performance comparison should be conducted with care, because deep learning approaches augment initial data by applying in-plane rotations, horizontal flips or reconstruct symmetric faces. In absence of data augmentation, deep-learning methods such as the one introduced by Lopes et al. \cite{lopes2017facial} obtains only 86.67\%. The same method obtains 96.76\% with data augmentation. Handcrafted approaches consider only the initial data. The performances achieved using only the initial data are well positioned with regard to the augmented settings.


\paragraph{Experiments on Oulu-CASIA}

Table \ref{stateCASIA} compares the performance of our method with major state-of-the-art approaches on Oulu-CASIA dataset under normal illumination and near infrared settings. The majority of approaches, evaluated on Oulu-CASIA dataset, takes into account only the data under normal illumination conditions (VL). Some approaches \cite{jeni2012robust,zhao2011facial} report performances on near infrared (NI) sequences to test their method in different light settings. Under various light settings available in the dataset, our method achieves better results than handcrafted approaches \cite{zhao2007dynamic,zhao2017facial,liu2016learning} and is competitive with regard to recent deep learning approaches \cite{jung2015joint,zhang2017facial,ding2017facenet2expnet}. The performances obtained in the near infrared domain outperform other approaches (81.88\%). According to the results, the combination of motion and geometric features clearly improves the performances (84.58\%) in the VL setting and our method obtains competitive performances. Still, under NI settings, LMP features perform the best due to robustness to poor landmarks detection.

\begin{table}[!h]
\renewcommand{\arraystretch}{1.3}
\caption{Performance comparison on Oulu-CASIA dataset \textit{(* Data augmentation / Deep learning)}.}
\label{stateCASIA}
\centering
\fontsize{7}{5}\selectfont
\begin{tabular}{lccccc}
\hline
\multicolumn{1}{c}{Method} & Measure & Seq. & Classes & VL-Acc(\%) & NI-Acc(\%) \\
\hline
LBP-TOP \cite{zhao2007dynamic} & 10-fold & 480 & 6 & 68.13\% & - \\
CLM + 3D Landmarks \cite{jeni2012robust} & LOSO & 480 & 6 & 72.31\% & 71.59\% \\
AdaLBP \cite{zhao2011facial} & 10-fold & 480 & 6 & 73.54\% & 72.09\% \\
LBP-TOP + Gabor \cite{zhao2017facial} & 10-fold & 480 & 6 & 74.37\% & - \\
Dis-ExpLet \cite{liu2016learning} & 10-fold & 480 & 6 & 79.00\% & - \\
DTAGN (joint) \cite{jung2015joint} & 10-fold & augm. & 6 & * 81.46\% & - \\
PHRNN-MSCNN \cite{zhang2017facial} & 10-fold & augm. & 6 & * 86.25\% & - \\
FN2EN \cite{ding2017facenet2expnet} & 10-fold & augm. & 6 & \underline{* 87.71\%} & - \\
\textbf{Local Motion Patterns (LMP)} & \textbf{10-fold} & \textbf{480} & \textbf{6} & \textbf{75.13\%} & \underline{\textbf{81.88\%}} \\
\textbf{LMP + Geometric features} & \textbf{10-fold} & \textbf{480} & \textbf{6} & \textbf{84.58\%} & \textbf{81.49\%} \\
\hline
\end{tabular}
\end{table}

\paragraph{Experiments on MMI}

Table \ref{stateMMI} compares the performance of recent state-of-the-art approaches on MMI dataset. We have selected only the activation sequence (e.g. neutral to apex) for 205 sequences. The combination of motion and geometric features clearly improves the performances (78.26\%) and out stands handcrafted approaches \cite{zhao2007dynamic,zhao2017facial,zhong2012learning,liu2016learning}. Compared to deep learning approaches \cite{zhang2017facial,mollahosseini2016going,jung2015joint} the results obtained are competitive considering that only the initial data was used for training.

\begin{table}[!h]
\renewcommand{\arraystretch}{1.3}
\caption{Performance comparison on MMI dataset \textit{(* Data augmentation / Deep learning)}.}
\label{stateMMI}
\centering
\fontsize{7}{5}\selectfont
\begin{tabular}{lcccc}
\hline
\multicolumn{1}{c}{Method} & Measure & Seq. & Classes & Acc(\%) \\
\hline
LBP-TOP \cite{zhao2007dynamic} & 10-fold & 205 & 6 & 59.51\%  \\
LBP-TOP + Gabor \cite{zhao2017facial} & 10-fold & 203 & 6 & 71.92\%  \\
Dis-ExpLet \cite{liu2016learning} & 10-fold & 205 & 6 & 77.60\%  \\
DTAGN (joint) \cite{jung2015joint} & 10-fold & augmented & 6 & * 70.24\% \\
Deep Neural Networks \cite{mollahosseini2016going} & 5-fold & augmented & 6 & * 77.60\% \\
PHRNN-MSCNN \cite{zhang2017facial} & 10-fold & augmented & 6 & * \underline{81.18\%} \\
\textbf{Local Motion Patterns (LMP)} & \textbf{10-fold} & \textbf{205} & \textbf{6} & \textbf{74.40\%} \\
\textbf{LMP + Geometric features} & \textbf{10-fold} & \textbf{205} & \textbf{6} & \textbf{78.26\%} \\
\hline
\end{tabular}
\end{table}

In the next section, we synthesize the results and we highlight our capacity to deal in a unified manner with the various challenges brought by micro and macro expressions.

\subsection{Micro and macro expression evaluation synthesis}

Table \ref{stateCross} summarizes the most relevant comparison results with representative state-of-the-art approaches on micro and macro expressions. We obtain very good results under varying illumination condition for both micro and macro expression recognition. The combination of the motion and the geometric features gives good performances in presence of small head pose variations for macro expression recognition. And more importantly, our method has the singularity of dealing in a unified manner with both micro and macro expressions challenges.

\begin{table}[!h]
\renewcommand{\arraystretch}{1.3}
\caption{Performance synthesis on all datasets \textit{(* Data augmentation / Deep learning)}.}
\label{stateCross}
\centering
\fontsize{6}{5}\selectfont
\begin{tabular}{|l|c|c|c|c|c|c|c|c|}
\cline{2-9}
\multicolumn{1}{c|}{} & \multicolumn{4}{c|}{Micro expression} & \multicolumn{4}{c|}{Macro expression} \\ \hline
\multirow{2}*{Method} & \multirow{2}*{CASME II} & \multicolumn{3}{c|}{SMIC} & CK+ & \multicolumn{2}{c|}{CASIA} & \multirow{2}*{MMI} \\
 & & HS & VIS & NIR & 7 classes & VL & NI & \\ \hline
LBP-TOP \cite{zhao2007dynamic} & - & - & 52.11\% & - & 96.26\% & 68.13\% & - & 59.51\% \\ \hline
LBP-TOP + Gabor\cite{zhao2017facial} & - & - & - & - & 95.80\% & 74.37\% & - & 71.92\% \\ \hline
AdaLBP \cite{zhao2011facial} & - & - & - & - & - & 73.54\% & 72.09\% & - \\ \hline
Dis-ExpLet \cite{liu2016learning} & - & - & - & - & 95.10\% & 79.00\% & - & 77.60\% \\ \hline
HIGO + magnification \cite{li2015reading} & 67.21\% & \underline{68.29\% }& 81.69\% & 67.61\% & - & - & - & - \\ \hline
* CNN + LSTM \cite{kim2017multi} & 60.98\% & - & - & - & - & - & - & - \\ \hline
* Sel. Deep feat. (CNN) \cite{patel2016selective} & - & 53.60\% & 56.30\% & - & - & - & - & - \\ \hline
* CNN + LSTM \cite{breuer2017deep} & 59.47\% & - & - & - & \underline{98.62\%} & - & - & - \\ \hline
* PHRNN-MSCNN \cite{zhang2017facial} & - & - & - & - & 98.50\% & 86.25\% & - & \underline{81.18\%} \\ \hline
* FN2EN \cite{ding2017facenet2expnet} & - & - & - & - & - & \underline{87.71\%} & - & - \\ \hline
\textbf{Local Motion Pattern} & \underline{\textbf{70.20\%}} & \textbf{67.68\%} & \underline{\textbf{86.11\%}} & \underline{\textbf{80.56\%}} & \textbf{97.25\%} & \textbf{84.58\%} & \underline{\textbf{81.46\%}} & \textbf{78.26\%} \\ \hline
\end{tabular}
\end{table}




The parameters used to assess LMP performances on each dataset are given in Table \ref{stateParam}. LMP settings vary slightly depending on the dataset, and, most of the time, the variations are due to the acquisition conditions (distance to the camera, resolution, frame rates). The parameters values vary within the intervals defined in section \ref{sec:parameters}.

\begin{table}[!h]
\renewcommand{\arraystretch}{1.3}
\caption{Parameter settings used for assessing the best results on different datasets.}
\label{stateParam}
\centering
\fontsize{7}{5}\selectfont
\begin{tabular}{|c|c|c|c|c|c|c|c|c|c|}
\hline
\multicolumn{2}{|c|}{Datasets} & $\lambda$ & $\Delta$ & $\rho$ & E & M & V & $\beta$ & bin \\
\hline
\parbox[t]{2mm}{\multirow{4}{*}{\rotatebox[origin=c]{90}{Micro}}} & CASME II & 4 & 0.5 & 0.75 & 100 & 4 & 5 & 6 & 9 \\
\cline{2-10}
& SMIC-HS & 3 & 0.5 & 0.75 & 100 & 3 & 5 & 6 & 9 \\
\cline{2-10}
& SMIC-VIS & 5 & 0.5 & 0.75 & 100 & 4 & 5 & 3 & 9 \\
\cline{2-10}
& SMIC-NIR & 4 & 0.5 & 0.75 & 100 & 3 & 5 & 3 & 12 \\
\hline
\parbox[t]{2mm}{\multirow{4}{*}{\rotatebox[origin=c]{90}{Macro}}} & CK+ & 3 & 0.5 & 1 & 100 & 4 & 5 & 3 & 12 \\
\cline{2-10}
& MMI & 3 & 0.5 & 1 & 100 & 4 & 5 & 6 & 12 \\
\cline{2-10}
& CASIA-VL & 4 & 0.5 & 1 & 100 & 5 & 5 & 3 & 6 \\
\cline{2-10}
& CASIA-NI & 5 & 0.5 & 0.75 & 100 & 5 & 5 & 6 & 9 \\
\hline
\end{tabular}
\end{table}

Results obtained for micro and macro expression prove the efficiency and the robustness of our contribution, which stands as a good candidate for challenging contexts (e.g. variations in head movements, illumination, activation patterns and intensities).

\section{Conclusion}
\label{sec:conclusion}

The main contributions of our paper are articulated around three axes. The first one is an innovative local motion patterns feature that measures temporal physical phenomena related to skin elasticity of facial expression. The second one is the unified recognition approach of both macro and micro expressions. The spatio-temporal features, extracted from videos, encode motion propagation into local motion regions situated near expression epicenters. As motion is inherent to any facial expressions our method is naturally suitable to deal with all expressions that cause facial skin deformation. The third one is related to the exponential potentiality and suitability of our method to meet in-the-wild requirements. We obtain good performances in various illumination (near infrared and natural) conditions for both micro and macro expression recognition. 
The method outperforms micro expression state-of-the-art methods on CASME II (70.20\%) and SMIC-VIS (86.11\%). Furthermore, we obtain without any additional data augmentations 
competitive results for macro expression recognition (97.25\% for CK+, 84.58\% for Oulu-CASIA and 78.26\% for MMI). 

Although our contribution narrows the gap with in-the-wild settings, other challenges such as dynamic background, occlusion, non-frontal poses, important head movements are still to be addressed. For example, let us consider the challenge of expression recognition in presence of important head movements. Although dynamic texture approaches perform well when analyzing facial expression in near frontal view, recognition of dynamic textures in presence of head movements remains a challenging problem. Indeed, dynamic textures must be well segmented in space and time. However, we believe that the registration based on facial components or shape are not adapted to dynamic approaches. Such registrations cause facial deformations and induce noisy motion. We believe that suitable relationship between motion representation and registration is the key for expression recognition in presence of head movements.

\section*{Acknowledge}
The authors wish to thank Mr. José Mennesson and Mr. Zhongfei (Mark) Zhang for their valuable discussions.


\end{document}